\documentclass[letterpaper, 10pt, conference]{ieeeconf}  

\pdfoutput=1
\let\labelindent\relax

\usepackage{graphicx}
\usepackage{amsmath}
\usepackage{multirow}
\usepackage{enumitem}
\usepackage{algpseudocode}
\usepackage{algorithm}
\usepackage{dsfont}
\usepackage{todonotes}

\newcommand{\ignorethis}[1]{}

\IEEEoverridecommandlockouts                              

\title{\LARGE \bf
Learning Agile Locomotion via Adversarial Training
}

\author{Yujin Tang$^{1,2}$, Jie Tan$^{3}$ and Tatsuya Harada$^{1}$
\thanks{$^{1}$Department of Mechano-Informatics, Graduate School of Information Science and Technology, The University of Tokyo.}%
\thanks{$^{2}$Google Brain, Tokyo, Japan.}%
\thanks{$^{3}$Google Brain, Mountain View, United States.}%
\thanks{Primary contact: \tt\small yujintang@google.com}
}

\begin{document}

\maketitle
\thispagestyle{empty}
\pagestyle{empty}

\begin{abstract}

Developing controllers for agile locomotion is a long-standing challenge for legged robots.
Reinforcement learning (RL) and Evolution Strategy (ES) hold the promise of automating the design process of such controllers. However, dedicated and careful human effort is required to design training environments to promote agility.
In this paper, we present a multi-agent learning system, in which a quadruped robot (protagonist) learns to chase another robot (adversary) while the latter learns to escape.
We find that this adversarial training process not only encourages agile behaviors but also effectively alleviates the laborious environment design effort.
In contrast to prior works that used only one adversary, we find that training an ensemble of adversaries, each of which specializes in a different escaping strategy, is essential for the protagonist to master agility.
Through extensive experiments, we show that the locomotion controller learned with adversarial training significantly outperforms carefully designed baselines.

\end{abstract}

\section{INTRODUCTION}

Despite decades of research on locomotion, legged robots have not yet demonstrated comparable agility to their animal counterparts.
Agile locomotion requires coordinated control of legs, precise manipulation of contact forces and intricate balance control.
The underlying control principles are still largely unknown to us.
Prior research has explored different gaits \cite{hyun2014high,hyun2016implementation,park2017high}, actuated tails \cite{briggs2012tails,patel2014rapid} and flexible spines \cite{khoramshahi2013piecewise,eckert2015comparing} to reproduce the agility of legged robots.
However, it is extremely challenging to rely on prior knowledge or manual tuning to design controllers for agile locomotion. 

\begin{figure}[h]
\centering
\includegraphics[scale=0.1]{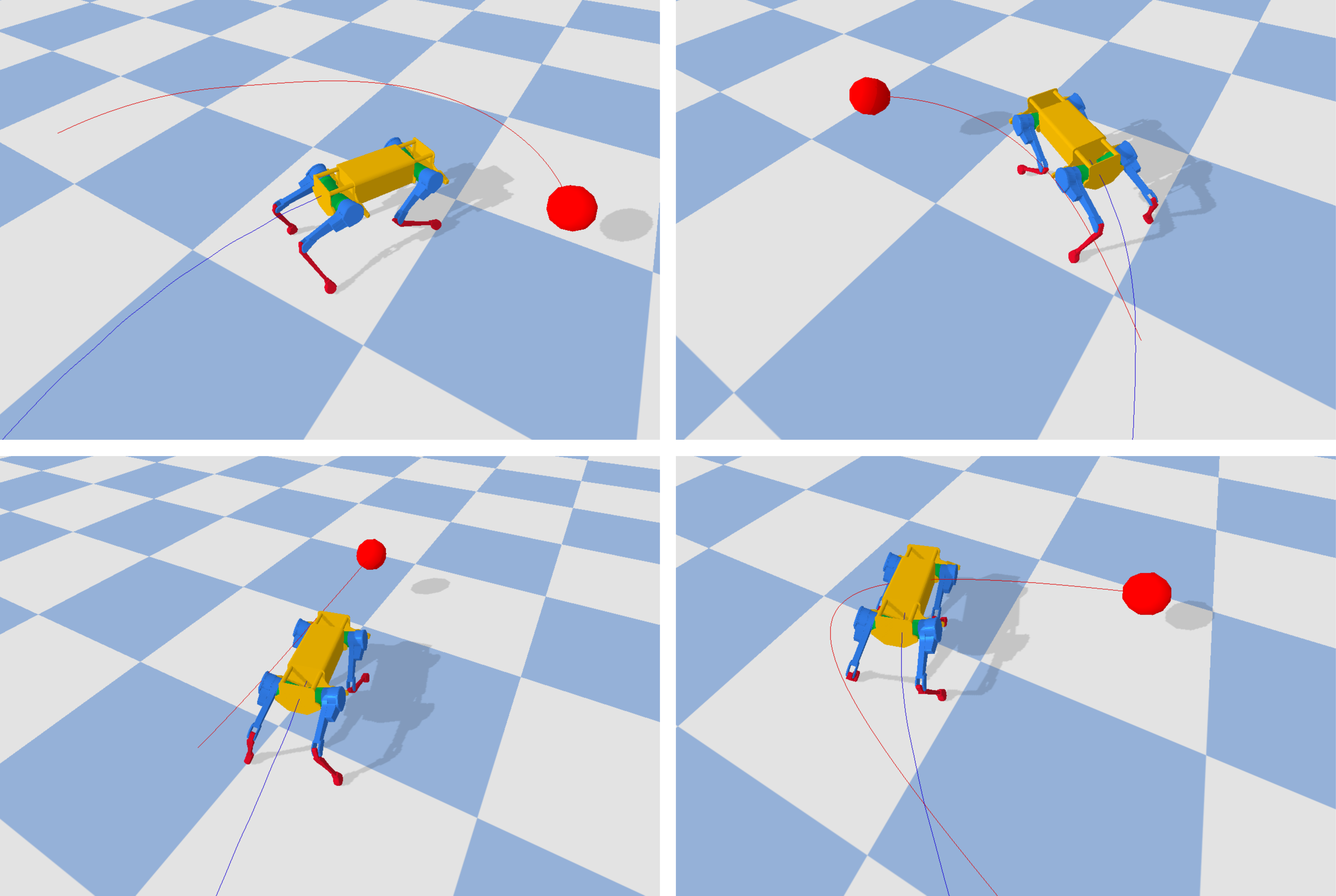}
\caption{
Sample episodes of chase and escape. The quadruped robot is the chaser and the red dot-bot is the escapee; the blue and red lines are their trajectories. In our experiments, some adversarial agents learned advanced escaping maneuvers such as induce the quadruped robot to come close, dodge and halt so that the quadruped robot runs past them.
}
\label{fig:env_samples}
\end{figure}

Reinforcement Learning (RL) \cite{schulman2017proximal,salimans2017evolution} and Evolution Strategy (ES) \cite{salimans2017evolution,mania2018simple} hold the promise of automating the design process of locomotion controllers \cite{47151,47598,hwangbo2019learning}.
Can we leverage these techniques to learn agility?
The main challenge is to design an environment and a reward function, with which agile locomotion can emerge after training.
Common practices for learning locomotion include using a reward function that encourages higher running speed, or a random initialization of target locations so that the robot learns to steer. However, neither setup leads to truly agile motion. 

Our goal in this study is to develop a learning system with which robots can learn agility without dedicated and careful human design efforts.
Inspired by pursuit-evasion behaviors between predators and preys in nature \cite{moore2015outrun}, we devise a multi-agent learning system \cite{4445757, OpenAI_dota, alphastarblog, Baker2020Emergent}, in which two agents learn to compete with each other. Adversarial training, or self-play has been extensively used in the field of games \cite{44806,DBLP:journals/corr/abs-1712-01815,alphastarblog} and robotics \cite{stone1998towards,45924,DBLP:journals/corr/abs-1903-00636,pmlr-v70-pinto17a}.
In our training scheme, one robot, the protagonist, learns to chase the other robot, the adversary, while the latter learns to escape (see Figure \ref{fig:env_samples}).
Both robots co-evolve their strategies. We find that many state-of-the-art multi-agent RL algorithms, such as MADDPG \cite{DBLP:journals/corr/LoweWTHAM17} and MATD3 \cite{ackermann2019reducing}, do not work in our settings. Learning locomotion and multi-agent interactions simultaneously presents significant challenges to these actor-critic based techniques. We decouple the problem, and train chasing and escaping behaviors iteratively.
While both RL and ES can learn locomotion controllers, we choose an ES approach in this paper due to its effectiveness on locomotion tasks~\cite{wu2010terrain,wang2010optimizing,tan2016simulation}.
Additionally, we augment Covariance Matrix Adaptation-Evolution Strategy (CMA-ES)~\cite{Hansen2006} to the multi-agent setting. Furthermore, to prevent the protagonist from overfitting to the running pattern of the adversary, we train an ensemble of adversaries, each of which specializes in a different escaping strategy. 

Our principal contribution is an adversarial learning algorithm that promotes agile locomotion behaviors.
To the best of our knowledge, it is the first algorithm that applies multi-agent learning to acquire agility in locomotion tasks.
We evaluate our method on a simulated quadruped robot.
After merely three generations of adversarial training, agile locomotion gaits emerge automatically. Please watch the accompanying video for the learned gaits. 
We perform comprehensive evaluations, which show that our method significantly outperforms the baselines without adversarial training, as well as the state-of-the-art multi-agent RL algorithms.

\section{Background and Related Work}

\subsection{Markov Decision Process}
We formulate locomotion control as a Markov Decision Process (MDP) and solve it using an RL or ES algorithm.
An MDP is a tuple $(S, A, r, D, P_{sas'}, \gamma)$, where $S$ is the state space; $A$ is the action space; $r$ is the reward function; $D$ is the distribution of initial states $s_0$; $P_{sas'}$ is the transition probability; and $\gamma \in [0, 1]$ is the discount factor.
A learning algorithm then optimizes a policy $\pi: S \mapsto A$ that maximizes the expected accumulated rewards $R_\pi(s_0)=\sum_t \gamma^t r(s_t)$:
\begin{displaymath}
\pi^*=\arg\max E_{s_0\sim D} [R_\pi(s_0)]
\end{displaymath}
In this paper, we don't use discount factor: $\gamma = 1$, and we use CMA-ES to solve the MDP (described in the next section).

\subsection{Covariance Matrix Adaptation-Evolution Strategy}
Covariance Matrix Adaptation-Evolution Strategy (CMA-ES)~\cite{Hansen2006} is a stochastic and derivative-free method for non-convex optimization. In the context of RL, CMA-ES samples a Gaussian distribution in the policy parameter space to generate a population. Each policy in the population is evaluated using the accumulated reward after rollouts. CMA-ES ranks the individual policies in the population and updates the underlying distribution accordingly.
This iterative optimization process continues until convergence or a user-specified number of iterations is reached.

\subsection{Multi-agent Reinforcement Learning}
\ignorethis{
Multi-agent reinforcement learning (MARL) addresses the sequential decision-making problem of multiple autonomous agents that operate in a common environment, each of which aims to optimize its own long-term return by interacting with the environment and other agents~\cite{4445757}.
Adversarial training can be formulated as an MARL problem, and in our settings, we have two agents (the chaser and the escapee) that compete with each other.

Traditional RL approaches such as Q-Learning or policy gradient are poorly suited to multi-agent environments~\cite{DBLP:journals/corr/LoweWTHAM17}.
Because all agents' policies are simultaneously evolving, the MDP assumption can be violated and the environment becomes non-stationary from the perspective of any individual agent and presents learning stability challenges.
It therefore prevents the straightforward application of experience replay which is crucial for stabilizing deep Q-learning.
On the other hand, policy gradient methods usually exhibit very high variance when coordination of multiple agents is required.
Addressing these, MADDPG~\cite{DBLP:journals/corr/LoweWTHAM17} and MATD3~\cite{ackermann2019reducing} proposed to extend DDPG~\cite{pmlr-v32-silver14} and TD3~\cite{DBLP:journals/corr/abs-1802-09477} to multi-agent settings, where the actors remain independent for each agent but the critic is centralized and evaluates for an individual agent based on global information.
}
Many robotic applications involve multiple agents that collaborate or compete~\cite{DBLP:journals/corr/PengYWYTLW17, OpenAI_dota, 10.5555/2900929.2901013, DBLP:journals/corr/FoersterAFW16a, NIPS2016_6398, DBLP:journals/corr/MordatchA17}.
In these applications, an extra layer of complexity arises. The co-evolution of agents invalidates the MDP assumption of the environment, causing the environment to be non-stationary from any individual agent's perspective~\cite{DBLP:journals/corr/LoweWTHAM17}.
Multi-agent reinforcement learning (MARL)~\cite{10.5555/1483085, 4445757} aims to address this difficulty.

The conventional actor-critic framework~\cite{NIPS1999_1786, pmlr-v32-silver14, DBLP:journals/corr/abs-1802-09477} has been extended to a multi-agent setting using centralized training and decentralized execution~\cite{DBLP:journals/corr/LoweWTHAM17, DBLP:journals/corr/abs-1802-09477, DBLP:journals/corr/FoersterFANW17}. It is challenging to apply these techniques to solve our problem because learning locomotion is already hard in the actor-critic framework, and learning multi-agent interactions on top of it greatly magnifies the difficulty.

Encouraged by the success of AlphaZero~\cite{DBLP:journals/corr/abs-1712-01815}, several recent works have investigated another paradigm of MARL, self-play, in more complex environments~\cite{DBLP:journals/corr/abs-1710-03748, Baker2020Emergent}.
Our method falls into this category. It is similar to autocurricula~\cite{Baker2020Emergent} in the sense that the adversaries learn from the weakness of the protagonist in each generation and present new challenges in the next generation.
Although it is possible to train the protagonist and the adversary simultaneously ~\cite{Baker2020Emergent}, we find that significant tuning effort is required so that a neural network policy has sufficient capacity to learn and maintain a variety of chasing/escaping skills. In this paper, we take a simpler approach by maintaining an ensemble of adversaries and adding new agents to it iteratively. The ensemble evolves gradually, which stablizes the protagonist's learning process. We show empirically that this ensemble formulation is crucial because it effectively prevents overfitting and catastrophic forgetting.


\section{Adversarial Training for Agile Locomotion}

\subsection{Adversarial Training}

To train agile locomotion, we designed a multi-agent environment wherein a quadruped robot, the protagonist, learns to chase another robot, the adversary, while the latter learns to escape.
The adversary robot is said to be caught if its distance to the protagonist is less than a predefined threshold $d_{min}$.
We train the protagonist and the adversaries in an iterative process. Algorithm~\ref{alg:alg01} shows the training process.
In the rest of the paper, we use the terms ``protagonist'' and ``chaser'', ``adversary'' and ``escapee'', interchangeably.

\begin{algorithm}[h]
  \caption{Adversarial Training for Agile Locomotion}
  \label{alg:alg01}
  \begin{algorithmic}[1]
    \Function{AdversarialTraining}{$\pi_{p}$}
      \State $\pi_a^0 \gets \textsc{StaticInitialAdversary}()$
      \State $\Pi_a \gets \{\pi_a^0\}$ \Comment{Initialize the adversary ensemble.}
      \For  {$i=1$ to $N$}  \Comment{Iterate for N generations.}
        \State $\pi_{p} \gets \textsc{LearnToChase}(\pi_{p}, \Pi_a)$
        \State $\Pi^{\prime}_a \gets \textsc{LearnToEscape}(\pi_{p})$
        \State $\Pi_{a} \gets \Pi_{a} \cup \Pi^{\prime}_a$
      \EndFor
      \State \textbf{Return} $\pi_p$
    \EndFunction
    
    \State
    \Function{LearnToChase}{$\pi_{p}, \Pi_a$}
      \For {$i=1$ to $N_{p}$} \Comment{Run $N_p$ CMA iterations.}
        \State $r = 0$  \Comment{Initialize roll-out reward.}
        \For {$k=1$ to $K$} \Comment{Test on K adversaries.}
          \State $j \gets \textsc{SampleAdversaryIndex}()$
          \State $\pi_a \gets \Pi_a[j]$
          \State $r \gets r + \textsc{DoRollout}(\pi_{p}, \pi_a)$
        \EndFor
      \State $\pi_p \gets \textsc{OptimizePolicy}(r / K, \pi_p)$
      \EndFor
      \State \textbf{Return} $\pi_p$
    \EndFunction
    
    \State
    \Function{LearnToEscape}{$\pi_{p}$}
    \State $\Pi^{\prime}_a \gets \emptyset$
    \For {$k=1$ to $K$} \Comment{Train K adversaries.}
      \State $\pi_{a}^{k} \gets \textsc{RandomInitialWeights}()$
      \For {$i=1$ to $N_a$} \Comment{Run $N_a$ CMA iterations.}
        \State $r \gets \textsc{DoRollout}(\pi_p, \pi_a^k)$
        \State $\pi_a^k \gets \textsc{OptimizePolicy}(r, \pi_a^k)$
      \EndFor
      \State $\Pi^{\prime}_a \gets \Pi^{\prime}_a \cup \{\pi_a^k\}$
    \EndFor
    \State \textbf{Return} $\Pi^{\prime}_a$
    \EndFunction
  \end{algorithmic}
  \label{alg:overview}
\end{algorithm}

In general, we train the protagonist and the adversary ensemble iteratively in an outer loop wherein the policy optimizations for each are carried out in an inner loop.
To avoid ambiguity, we refer to each repetition in the outer loop as ``generation'' and that in the inner loop as ``iteration''.
In the first generation, we initialize the adversary ensemble to be $\Pi_a = \{\pi_a^0\}$, where $\pi_a^0$ is a simple policy that the escapee remains still (Line 2-3).
Next, we train both the protagonist and the adversary ensemble for $N$ generations.
In the $i$th generation, we first train the protagonist policy $\pi_p$ to play against the adversary policies selected from $\Pi_a$ (Line 5), then we train multiple adversary policies $\pi_a^{k}$ to find counter strategies against $\pi_p$ (Line 6).
We augment the adversary ensemble $\Pi_a$ with these newly learned adversaries (Line 7).
Both the protagonist and the adversarial policies are feed-forward neural networks, and they share the same architecture.
We use CMA-ES to train both the protagonist and the adversarial policies in $\textsc{OptimizePolicy}$ (Lines 20 and 31).
At the beginning of the $i$th generation, while the protagonist policy is warm-started from the last generation, adversaries are always initialized with random weights to encourage behavior diversity.
Note that the training of the $K$ adversaries in each generation are independent and can be carried out in parallel.
We defer the details of protagonist and adversary training to Sections~\ref{sec:learn_to_chase} and~\ref{sec:learn_to_escape}.

A key difference between our method and other adversarial training works in robotics~\cite{DBLP:journals/corr/abs-1903-00636,pmlr-v70-pinto17a} is that we train the protagonist from an ensemble of adversaries.
We find that training using a single adversary can lead to poor performance due to overfitting and catastrophic forgetting \cite{Chen:2016:LML:3086758}.
Intuitively, in the context of chase and escape, suppose the adversarial policy $\pi_a$ in the $i$th generation learned a sudden maneuver to the right side of the chaser, causing the chaser to fall.
The chaser's policy $\pi_p$ in the next generation will learn to counter this trick, for example, by always leaning its body towards the right, a behavior that is specific to this escaping strategy.
An adversary in the next generation can quickly spot this behavior and utilize it as an advantage by escaping to the left side. Moreover, though the protagonist policy in the next generation may again learn to counteract, it will forget what it has learned before, which is known as catastrophic forgetting.
Learning from an ensemble of adversaries solve both problems because the protagonist is constantly exposed to a wide variety of counter strategies. Please refer to Section \ref{sec:adversary_ensemble} for detailed evaluations on this.

\subsection{Learn to Chase}\label{sec:learn_to_chase}

We focus on legged robots in this paper wherein the chaser needs to learn the low level control, gaits, balance, steering, as well as decision making to quickly intercept the escapee. We formulate an MDP to learn this task. The MDP state space includes the robot's joint angles and velocities as well as the escapee's relative position. The action space includes the desired joint angles for all legs.

We reward the chaser based on the following four criteria:
(i) running toward the escapee robot and catch it as quickly as possible;
(ii) steering its heading towards the escapee;
(iii) keeping its pose close to a reference;
(iv) using a symmetric running style.
Therefore, we define the reward function as the following:
\begin{align}
r_t^{chaser} =&~~~e^{-|\theta_t|} (d_{t-1} - d_t)\nonumber\\
& + w_1\mathds{1}{(d_t \le d_{min})} \nonumber\\ 
& -w_2 \|\pi_p(s_t)-\bar{\mathbf{q}}\| \nonumber \\
& -w_3 \|\pi_p(s_t) - \Psi_a\Big{(} \pi_p\big{(} \Psi_s(s_t) \big{)} \Big{)}\|  \label{eq:reward}
\end{align}
The first term measures the distance changes between the chaser and the escapee in two adjacent time steps, discounted by the angle $\theta$ between the chaser's heading and the direction towards the escapee.
The second term is a bonus when the chaser successfully catches the escapee, where $\mathds{1}$ is an indicator function. The last two terms regulate the running style. The third term penalizes large pose deviations from a reference pose and the last term encourages the chaser to develop symmetric motions between the left and the right limbs. $\Psi_s$ and $\Psi_a$ mirror the state and the action along the sagittal plane at the center of the robot's body, same as those in Yu et al. \cite{yu2018learning}. $w_1$, $w_2$ and $w_3$ are the weights.

\begin{figure}[t]
\centering
\includegraphics[scale=0.30]{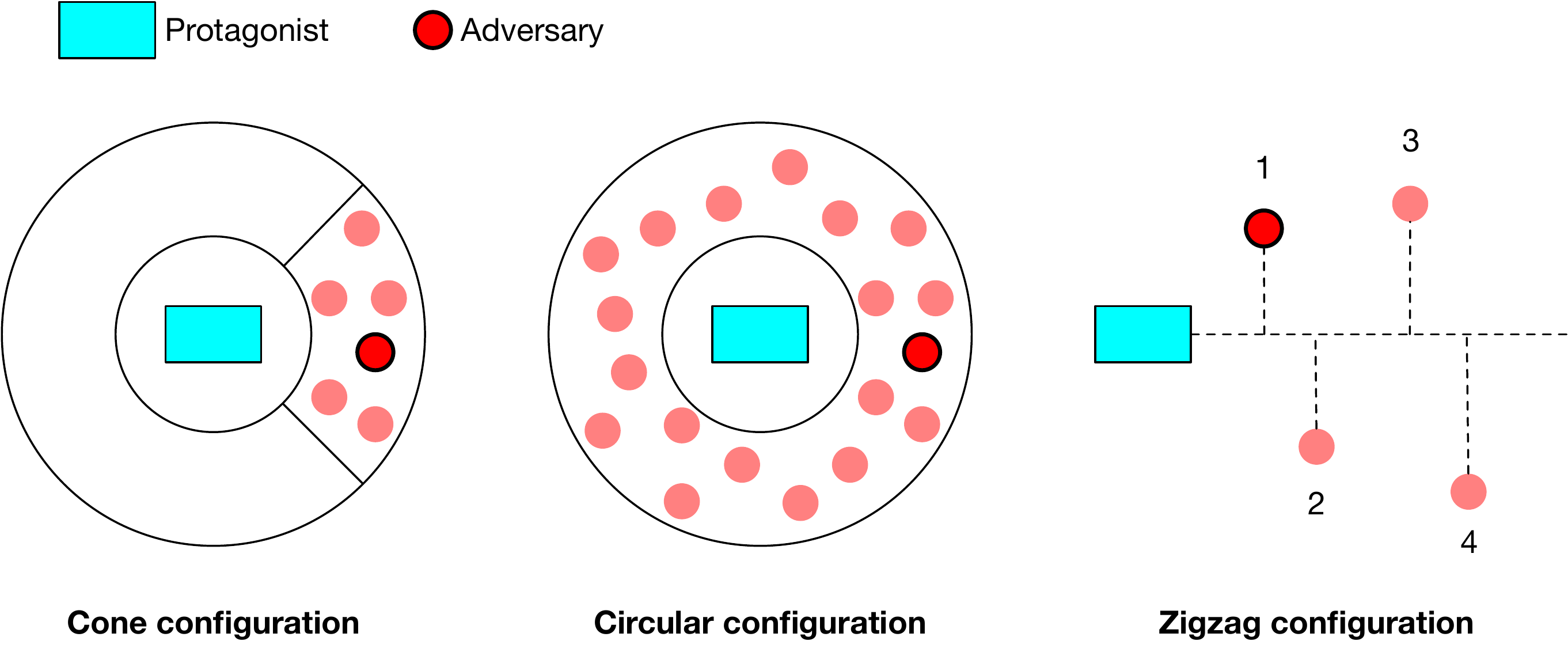}
\caption{Illustration of the three manually designed configurations from which adversary positions are sampled. The protagonist is facing right in the figure; the red dots with solid line boundary are the current sampled adversary position; light red dots are other possible sampling positions. In the zigzag configuration, we sample the vertices of a zigzag pattern before an episode begins and return them in the order depicted in the figure.}
\label{fig:init_placement}
\end{figure}

In each roll-out, we initialize the starting position of the escapee robot randomly in a partial circular ring with predefined angle and inner/outer radius in front of the chaser (Cone Configuration in Figure~\ref{fig:init_placement}).
Besides the cone configuration, we have also designed other escapee robot initialization configurations and trained controllers from them as our baselines in Section~\ref{sec:experiment}.
Each roll-out terminates when the chaser robot falls or after 2000 time steps, which is equivalent to 4 seconds.
If the chaser catches the escapee ($d_t \le d_{min}$), we sample a new adversary position relative to the current protagonist's position and orientation, using the same cone configuration, and continue the current episode.

\subsection{Learn to Escape}\label{sec:learn_to_escape}

Similar to the chaser, we formulate an MDP for the escaping task: to run away from the chaser. The escapee robot, besides its own states, also observes the chaser robot's base position and orientation in its local coordinate.
In our experiments, we use simple dot-bots (see Figure~\ref{fig:env_samples}) that are controlled kinematically as escapees, which makes it easier to learn diverse and effective escaping strategies. Refer to Section~\ref{sec:robots} for more details.

In contrast to a common practice in adversarial training, simply negating the protagonist's reward and assign it to the adversary does not work in our case. Negative rewards, combining with early termination, would make the escapee learn to suicide by running towards the chaser, to minimize the accumulation of negative rewards.
We therefore design a different reward for escapees, which only requires the escapee robot to stay far away from the chaser:
\begin{equation*}
r_t^{escapee} = d_{t} - d_{t-1}
\end{equation*}
where $d_{t}$ and $d_{t-1}$ are the distances between the chaser and the escapee in the current and previous time steps.

In addition to the reward design, there are two key differences between training the chaser and the escapees. First, in the chasing task, the distance threshold for catching is fixed: $d_{min}=0.5$. In the escaping task, we randomly perturb it: $d_{min} \in (0.5, 1.0]$ for each escapee to increase the diversity of escaping behaviors.
Intuitively, a small $d_{min}$ allows the escapee to stay close to the chaser, and develop sudden quick movements to dodge, while large $d_{min}$ would encourage the escapee to use large circular trajectories to stay away from the chaser.
Second, we only terminate the episode when the maximum number of steps has reached. We do not early terminate the episode when the chaser falls, because in this case, the escapee can accumulate more rewards and thus reinforce this effective strategy that induces the chaser to fall.

\section{Experiments}\label{sec:experiment} 

In this section, we evaluate our method on a simulated quadruped robot.
We design two sets of baselines. In the first set, we compare our method with two state-of-the-art actor-critic-based multi-agent RL algorithms:
MADDPG~\cite{DBLP:journals/corr/LoweWTHAM17} and MATD3~\cite{ackermann2019reducing}. In the second set of baselines, we aim to show that adversarial training learns highly agile behaviors that manually designed environment setups can hardly achieve.
Our goal with these evaluations is to answer the following questions:
\begin{enumerate}
    \item Does our method produce agile locomotion that the baselines cannot? 
    \item Does the learned locomotion controller generalize to unseen situations?
    \item Is training with the adversary ensemble essential?
\end{enumerate}

\begin{figure}[t]
\centering
\includegraphics[scale=0.28]{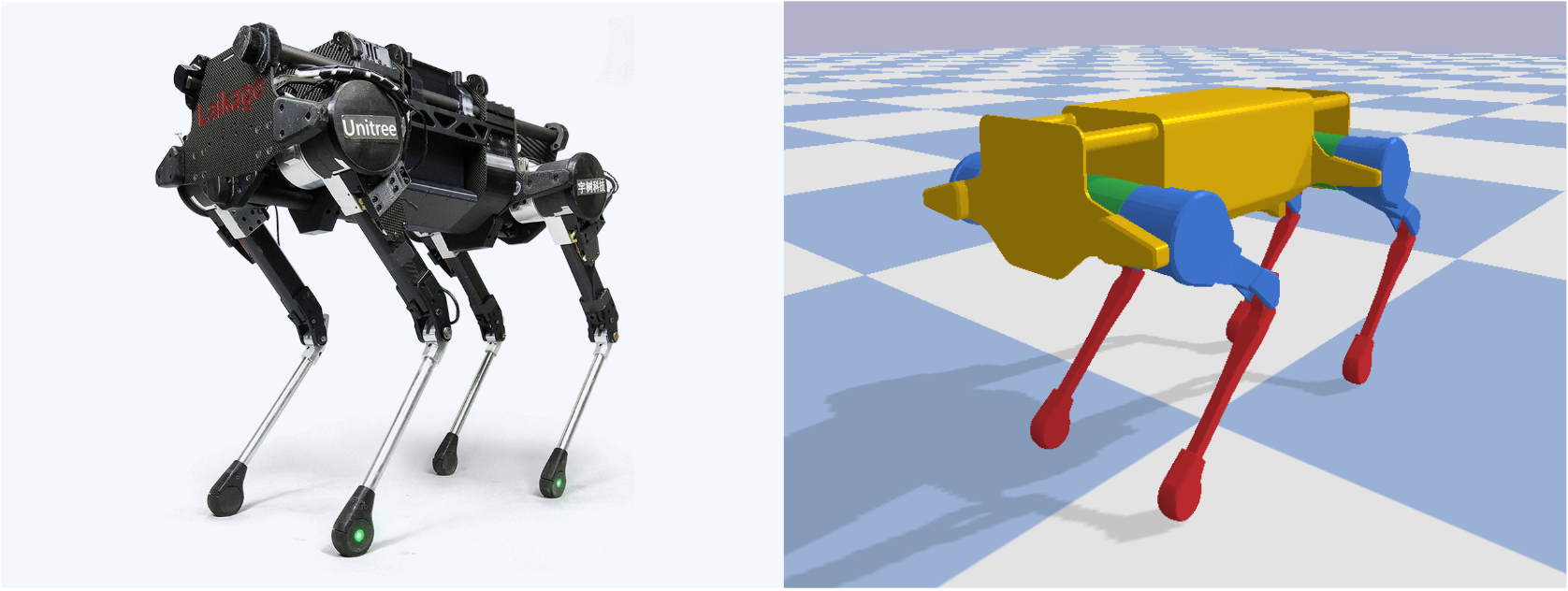}
\caption{We use Unitree Laikago as our chaser robot. (Left) The real Laikago robot. (Right) The simulated Laikago robot in a PyBullet physics simulator. The right figure shows the reference pose $\bar{\mathbf{q}}$ in the chaser's reward function.}
\label{fig:laikago}
\end{figure}

\subsection{Robots}\label{sec:robots}
We model Unitree's Laikago~\cite{laikago-intro} in the PyBullet simulation \cite{coumans2020} (see Figure~\ref{fig:laikago}).
While the simulation has access to all ground-truth states of the environment and the robots, we select a subset of them as our robots' MDP states. This subset is sufficient for us to learn locomotion controllers.

The MDP state space of the chaser robot is a 27D vector, including the robot's joint angles (12D), angular velocities (12D) and the relative position of the escapee in its local frame (3D) at every simulation step.
Laikago has 12 actuated degrees of freedom, and hence the action space of our chaser MDP is a vector of 12 desired motor angles.
When the robot receives a command, we use PD control to convert the motor angles to torques and apply them to the actuated joints. In our simulation, the gains are $K_p=180$ and $K_d=8$. 

For the escapee, we design a dot-bot (see Figure~\ref{fig:env_samples}). The dot-bot floats at a fixed height above the ground and can move and rotate in the horizontal plane at linear and angular speeds up to 2 m/s and 2 rad/s. We tuned the escapee's highest speed to be slightly lower than that of the chaser to avoid it from learning trivial skills, such as running at full speed in a straight line. The dot-bot is controlled with twist commands: desired forward and spinning speeds $(v, \omega)$, similar to many wheeled robots. The states of a dot-bot $(x, y, \theta)$ evolve based on the following equations:
\begin{align}
\theta_{t+1}&=\theta_t + \omega_t\Delta t \nonumber \\
x_{t+1} &= x_t + v\cos(\theta_t)\Delta t \nonumber \\
y_{t+1} &= y_t + v\sin(\theta_t)\Delta t \nonumber 
\end{align}
The escapee robot observes the relative position of the chaser in its local frame, therefore its state space is a 3D vector.
Since this simple robot does not need to learn locomotion and balance, it can focus on learning diverse and competitive escaping strategies, which in turn makes the chaser more agile.

\subsection{Training Details}
In all our experiments, we run the adversarial training for $N=3$ generations. In the first generation, the escapee is placed according to the zigzag configuration (see Figure \ref{fig:init_placement}) and remains still for the entire episode. In future generations, the escapee is initialized using the cone configuration and moves according to its own policy $\pi^{k}_{a}$. 
In each generation, we learn $K=8$ adversaries and randomly split them into training and testing sets (4 in each set). Only the escapees in the training sets are added into the adversary ensemble $\Pi_a$.
We collect 16 adversaries throughout the entire training. Both the training and testing sets have 8 adversaries.

For both the chaser and the escapees, we use the same fully-connected neural networks of two hidden layers with 64 units per layer, and \textsc{tanh} activations, to represent their policies. We train all of them using CMA-ES~\cite{hansen2019pycma}, with a population size of 256, and initial standard deviation of 0.1.
For the chaser, we train $N_p=1K, 1K, 2K$ iterations for the three generations respectively. For each escapee, we train for $N_a=200$ iterations per generation.

\subsection{Results}

\begin{figure}[t]
\centering
\includegraphics[scale=0.42]{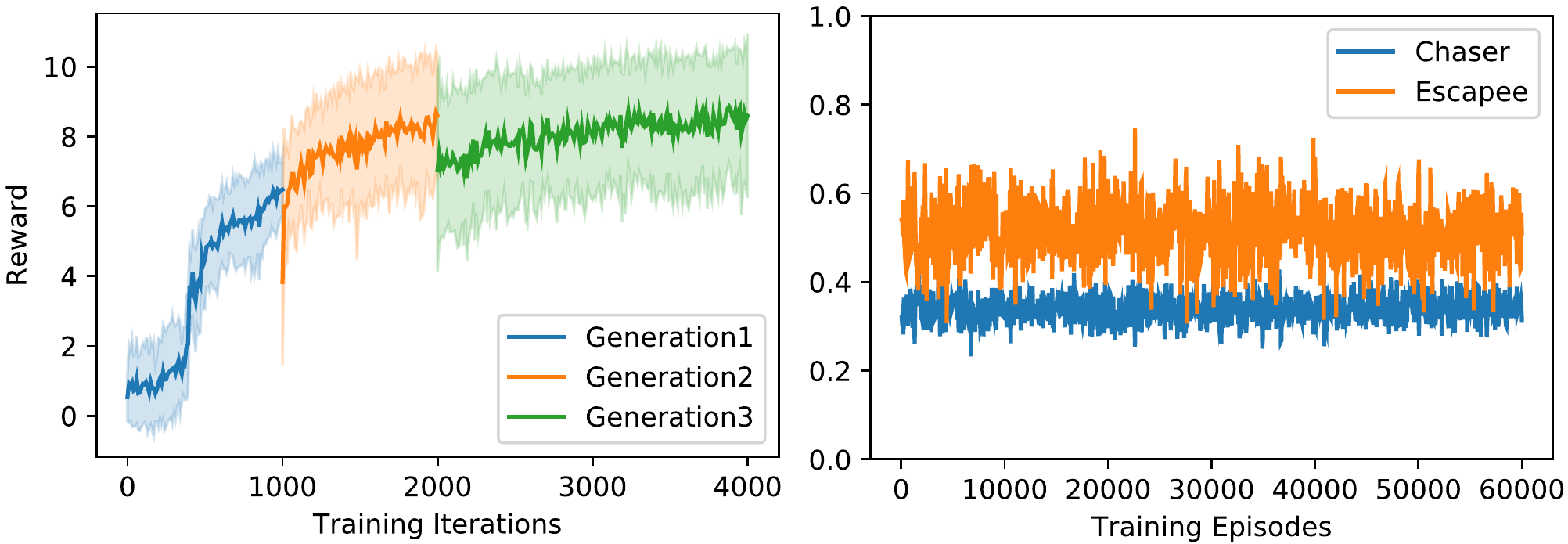}
\caption{Learning curves. (Left) We trained our controller with adversarial training for three generations. (Right) We also tried MADDPG~\cite{DBLP:journals/corr/LoweWTHAM17} and MATD3~\cite{ackermann2019reducing} to train the chaser and the escapee simultaneously, but they failed to learn successfully. The figure shows the learning curves from MADDPG, and MATD3's are similar.}
\label{fig:learning_curves}
\end{figure}

After training, the Laikago robot learns a symmetric gait that actuates its forelimbs and hindlimbs in turn, which resembles a bounding gait that is commonly seen when quadruped animals run at high speeds.
To achieve sharp turns, the robot elongates the stance phase of a forelimb so that it can be used as a pivot to rotate the entire body, which changes its heading direction rapidly. 

Figure~\ref{fig:learning_curves} (left) shows the chaser's learning curve from our adversarial training. The curve shows the average reward and the standard deviation over 100 episodes, each of which is to chase a randomly selected adversary.
In the first generation where the adversary is static (blue segment), the chaser learns basic locomotion skills.
The large amount of iterations needed to achieve a good reward proves that learning locomotion is a nontrivial task.
In the second generation (orange segment), the chaser needs to play against four escapees in addition to the initial static one.
We find interesting maneuvers that these adversaries learned, including running at full speed, in circles, and waiting for the chaser but suddenly dodging to the side to let it overshoot.
These strategies were effective initially, as indicated by the abrupt drop of the chaser's performance at the beginning of the orange segment.
However, the chaser quickly learns to counter these strategies, and the learning curve quickly rises back and climbs up.
In the last generation (green segment), four more newly-trained adversaries are added to the ensemble. They also learned different strategies, such as circling around with different curvatures.
Similar to the beginning of the second generation, we again observe an initial drop of performance, and then the learning progresses at a steady pace.

\subsection{Comparison with Actor-Critic MARL}
We apply two state-of-the-art MARL algorithms, MADDPG~\cite{DBLP:journals/corr/LoweWTHAM17} and MATD3~\cite{ackermann2019reducing}, to our problem.
We use the same network architecture in our proposed method and apply grid-search for hyper-parameters.
However, neither is successful in learning agile locomotion behaviors. The learning curves using MADDPG stay flat for both agents (see Figure~\ref{fig:learning_curves} right). The learning curves of MATD3 are similar but not shown. After 3 days of training (the same wall-clock time as our adversarial training), the chaser cannot even walk stably. We suspect that learning basic locomotion controllers and high-level pursuit-evasion behaviors simultaneously may be too challenging for the actor-critic multi-agent RL algorithms.

\subsection{Comparison with Manually Designed Environments without Adversarial Training}
We carefully design three baselines to promote agile behavior but without adversarial training: The dot-bot is placed strategically but does not learn to escape during training, see Figure~\ref{fig:init_placement}.
\begin{itemize}
    \item The cone configuration randomly places the escapee in a partial circular ring with predefined angle and inner/outer radius in front of the chaser. It is the same as how we initialize escapee's position in adversarial training. 
    \item The circular configuration is similar to the cone configuration, except that it allows positioning the adversary in the full circular ring. Placing the adversary behind the chaser would encourage it to learn sharp turns.
    \item The zigzag configuration samples a series of points that form a zigzag pattern when the episode starts. If an adversary is caught, the next point in the zigzag pattern is retrieved to re-place the adversary. Running according to the zigzag pattern requires the chaser to maintain balance when experiencing high lateral acceleration. This is a key metric for agility in prior studies \cite{wilson2013locomotion}.
\end{itemize}
We denote the policies trained using these baselines and our method as $\pi_{cone}$, $\pi_{circular}$, $\pi_{zigzag}$, and $\pi_{adversary}$ respectively.

We train four chasers using the three baselines as well as our adversarial training procedure, cross-validate them in all training environments and summarize the rewards (Equation \ref{eq:reward}) in Table~\ref{tab:test_in_train}. For example, the entry ($\pi_{cone}$, Adversary) in the table is the performance of the protagonist that was trained in the Cone environment and tested to chase adversaries trained using our method. For each entry, we run the policy in the corresponding environment for 100 episodes and report the mean reward.
We also normalize the rewards for each column by dividing the reward in the diagonal entry that evaluates the policy in the same training environment (e.g. testing $\pi_{cone}$ in the Cone environment).
The rightmost column shows the average normalized score for tests across all the environments. $\pi_{adversary}$ clearly dominates almost all the tests. This means that the chaser trained using our method is more agile when chasing adversaries in baseline environments than vice versa.
Therefore, our adversarial training procedure can learn more agile locomotion skills.

\ignorethis{All the baselines and the proposed method are successfully learned, as is demonstrated by the fact that each method's reward is among the tops in the corresponding training environment.
Compared with others, $\pi_{adversary}$ not only does well in its own but also keeps the rewards high in the baselines' training environments.
Among the three baselines, $\pi_{zigzag}$ gives the best performance,
but $\pi_{adversary}$ can still exceed its performance by 14\% in terms of the average reward.
We find this interesting as it bears some resemblance to human athletes training.
Zigzagging is a common exercise for agility, yet to ace competitors athletes need targeted training to overcome the problems specific to themselves.
Adversarial training provides opportunities for discovering these problems (by training an ensemble of adversaries) and targeted training (by playing against these adversaries).}

\begin{figure*}[!h]
\centering
\includegraphics[scale=0.35]{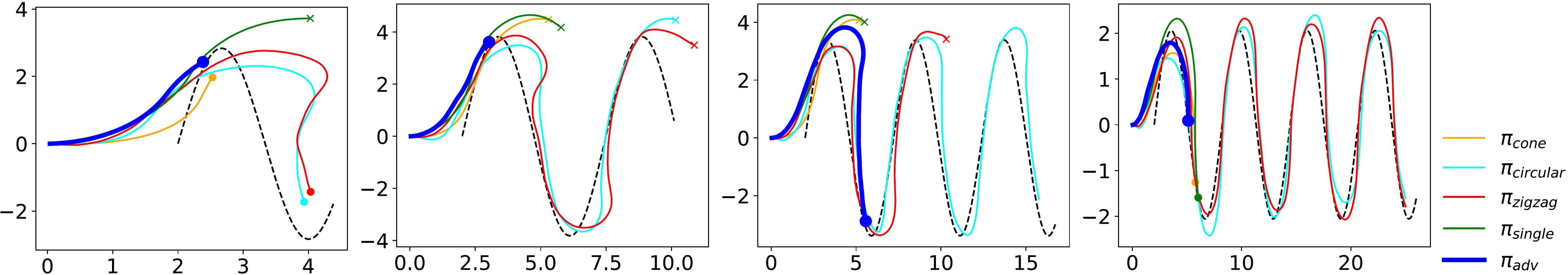}
\caption{Chaser's trajectories when chasing a target that follows sine curves with different amplitudes and frequencies. The target moves along the dotted curve.
A cross at the end of a trajectory indicates that the chaser has fallen or the target has escaped. A dot at the end means successfully catching the target at that position. Short trajectories ending with dots indicate the chaser catches the target early. The chaser trained using our method (blue trajectory) is able to catch the target much earlier than other baseline policies.}
\label{fig:sample_trajs}
\end{figure*}

\begin{table}[t]
\caption{\normalfont In addition to its own, each method is also tested in other methods' training environments. For each method and environment pair, we test for 100 episodes and we report the normalized mean reward. Average rewards from all environments are summarized in the rightmost column. Best scores in each column are highlighted in boldface.}
\label{tab:test_in_train}
\begin{tabular}{llllll}
\hline
\textbf{}            & {Cone} & {Circular} & {Zigzag} & {Adversary} & {Average} \\ \hline
$\pi_{cone}$         & 1.00          & 0.58              & 0.65            & 0.57               & 0.70         \\
$\pi_{circular}$     & 0.96          & 1.00              & 0.84            & 0.56               & 0.84         \\
$\pi_{zigzag}$       & 1.02          & 0.94              & \textbf{1.00}            & 0.65               & 0.90         \\ \hline
$\pi_{single}$       & 1.07          & 0.64              & 0.70            & 0.65               & 0.77         \\
$\pi_{adversary}$          & \textbf{1.15}          & \textbf{1.01}              & 0.96            & \textbf{1.00}               & \textbf{1.03}         \\ \hline
\end{tabular}
\end{table}

\subsection{Chasing an Escapee along Sine Trajectories}\label{sec:sine_test}
Studies \cite{wilson2013locomotion} have shown that the key to animals' agile locomotion, such as cheetah’s hunting runs, is to maintain balance under huge lateral acceleration when the animal is negotiating tight turns at high speed. For this reason, in this section, we evaluate agility of legged robots as the ability to follow trajectories with large curvatures at high speed.

To generate pursuit trajectories with different curvatures, we sample 100 sine curves with random amplitudes $A$ and frequencies $\omega$ that extends along the $x$-axis.
\begin{displaymath}
y = A\sin\big{(}\omega (x-2)\big{)}
\end{displaymath}
where $A \in [2, 4]$ and $\omega \in [0.1 \pi, \pi]$.
In this experiment, the chaser starts at the origin and the target starts 2 meters ahead of it.
The chaser needs to catch the target that travels along the sine curve at 2m/s within 10000 steps (20 seconds).
An episode ends early if the chaser falls or catches the target. The target is considered to have successfully escaped if its distance to the chaser is larger than 3 meters or the chaser cannot catch it before the episode ends.
Table~\ref{tab:sine_test} summarizes the test metrics.
In these tests, we measure the percentage that the chaser falls, catches the target, and the target successful escapes.
We also record the average distance between the chaser and the target during the chasing process (in meter), the chaser's speed (in meter/second) and the error of its heading direction (denoted as $\theta$ in the table, in radians).

Table~\ref{tab:sine_test} (top) clearly shows that our method significantly outperforms all other baselines. Our policy is more agile since it has the highest catch rate, and the lowest fall and escape rates.
Compared with baselines, our chaser can run at a higher speed, maintain a shorter distance to the target and align its heading direction closer towards the target.

Figure~\ref{fig:sample_trajs} shows four example trajectories.
In the first two cases, our policy can catch the target even before it enters the first turn. In contrast, the policies trained with baselines either fall or need longer time to chase.
In cases when the target enters the turns (the last two plots), our policy can tightly follow the trajectory and catch the target, while the baseline policies either lose balance or have to reduce the speed to negotiate the turn.
Clearly, our adversarial training scheme learns agile locomotion that cannot be acquired through the baseline environments that we carefully design. 

\begin{table}[]
\caption{\normalfont Results from chasing target along desired trajectories (top) and chasing unseen adversaries (bottom). Distance (m), speed (m/s) and orientation offset $\theta$ (radian) are mean values from each test episode and averaged over all tests. Smaller values are preferred for fall, escape, distance and $\theta$, and larger values are preferred for catch and speed. Best metrics in each column are highlighted in boldface.}
\label{tab:sine_test}
\begin{tabular}{lllllll}

\multicolumn{7}{l}{\textbf{Chasing along Trajectories}}         \\ \hline
                 & {Fall} & {Catch} & {Escape} & {Distance} & {Speed} & $\mathbf{\theta}$ \\ \hline
$\pi_{cone}$     & 11            & 61             & 28        & 1.72              & 2.24           & 0.43              \\
$\pi_{circular}$ & \textbf{0}             & 37             & 63           & 1.84              & 1.98           & 0.47              \\
$\pi_{zigzag}$   & 7             & 35             & 58         & 1.60              & 2.05           & 0.44              \\ \hline
$\pi_{single}$   & 3             & 65             & 32          & 1.56              & \textbf{2.45}           & 0.43              \\ \hline
$\pi_{adversary}$      & \textbf{0}             & \textbf{97}             & \textbf{3}          & \textbf{1.50}              & 2.40           & \textbf{0.29}              \\ \hline
\\
\multicolumn{7}{l}{\textbf{Chasing Unseen Adversaries}}         \\ 
\hline
                 & {Fall} & {Catch} & {Escape} & {Distance} & {Speed} & $\mathbf{\theta}$ \\ \hline
$\pi_{cone}$     & 1             & 30             & 69           & \textbf{1.92}              & 2.46           & 0.47              \\
$\pi_{circular}$ & \textbf{0}             & 13             & 87        & 2.20              & 2.11           & 0.27              \\
$\pi_{zigzag}$   & 5             & 16             & 79         & 2.05              & 2.18           & 0.27              \\ \hline
$\pi_{single}$   & \textbf{0}             & 51             & 49       & 2.13              & \textbf{2.50}           & 0.41              \\ \hline
$\pi_{adversary}$      & \textbf{0}             & \textbf{98}      & \textbf{2}              & 2.06              & 2.43           & \textbf{0.14}              \\ \hline
\end{tabular}
\end{table}

\subsection{Generalization to the Unseen Adversaries}\label{sec:unseen_adversaries}

In this experiment, we evaluate the generalization of the chaser policy: the chaser robot needs to chase the eight adversaries that were set aside in the test set.
We test each policy for 100 episodes, in each of which one of the eight adversaries is sampled and placed at a random  location in front of the chaser. We use the same metrics as in Section~\ref{sec:sine_test} and summarize the results in Table~\ref{tab:sine_test} (bottom).
It is clear that the controllers trained using our method generalizes better than the baselines.
When encountering the escaping strategies that were not seen during training, the protagonist can still catch the adversary 98\% of the time and achieve better metrics than the baselines.

\subsection{Adversary Ensemble}\label{sec:adversary_ensemble}

To demonstrate that the adversary ensemble is essential for our method, we compare it with an implementation that uses only a single adversary ($\pi_{single}$).
Training with a single adversary does not bring out agility. The results are even worse than baselines in some tests. For example, in Table~\ref{tab:test_in_train}, $\pi_{single}$'s performance is second-to-last.
Although in Table~\ref{tab:sine_test}, $\pi_{single}$'s catch rate tops the baselines, it is only about half of $\pi_{adversary}$'s.
In the experiments of chasing a target along sine trajectories, we notice that both $\pi_{single}$'s speed and heading error $\theta$ are high, which suggests that $\pi_{single}$ relies on fast speed to catch the target, yet is unable to make sharp turns if the target's trajectory starts to curve.
After close examination, we realize that $\pi_{single}$ is trained against an adversary whose strategy is the side-way dodge, which can be effectively countered by fast running. This shows that $\pi_{single}$ overfits to that single adversary's strategy and fails to generalize. Therefore, the adversary ensemble is an essential component of our algorithm.

\section{Conclusion}

In this paper, we devise a multi-agent learning system to acquire agile locomotion behaviors for legged robots. 
The core ideas of our method include an iterative adversarial training process and learning from the adversary ensemble. During training, one robot, the protagonist, learns to chase the other robot, the adversary, while the latter learns to escape.
In our experiments, after three generations of adversarial training, agile locomotion gaits emerge automatically.
Our method significantly outperforms the carefully designed baselines that do not have the adversarial component.
One limitation of our work is that the learned agile motion does not resemble the gaits of real animals.
We suspect that it is because our simple reward design does not capture all important factors from a biological point of view, such as energy efficiency. One possible solution is to pre-train the control policy by imitating animals \cite{RoboImitationPeng20} before the adversarial training. Another avenue of future research is to overcome the sim-to-real gap and deploy the learned agile controllers on the real robot. 

\section*{Acknowledgements}
This work was partially supported by JST AIP Acceleration Research Grant Number JPMJCR20U3, and partially supported by JSPS KAKENHI Grant Number JP19H01115.  


\bibliographystyle{IEEEtran}


\begin{thebibliography}{10}
\providecommand{\url}[1]{#1}
\csname url@rmstyle\endcsname
\providecommand{\newblock}{\relax}
\providecommand{\bibinfo}[2]{#2}
\providecommand\BIBentrySTDinterwordspacing{\spaceskip=0pt\relax}
\providecommand\BIBentryALTinterwordstretchfactor{4}
\providecommand\BIBentryALTinterwordspacing{\spaceskip=\fontdimen2\font plus
\BIBentryALTinterwordstretchfactor\fontdimen3\font minus
  \fontdimen4\font\relax}
\providecommand\BIBforeignlanguage[2]{{%
\expandafter\ifx\csname l@#1\endcsname\relax
\typeout{** WARNING: IEEEtran.bst: No hyphenation pattern has been}%
\typeout{** loaded for the language `#1'. Using the pattern for}%
\typeout{** the default language instead.}%
\else
\language=\csname l@#1\endcsname
\fi
#2}}

\bibitem{hyun2014high}
D.~J. Hyun, S.~Seok, J.~Lee, and S.~Kim, ``High speed trot-running:
  Implementation of a hierarchical controller using proprioceptive impedance
  control on the mit cheetah,'' \emph{The International Journal of Robotics
  Research}, vol.~33, no.~11, pp. 1417--1445, 2014.

\bibitem{hyun2016implementation}
D.~J. Hyun, J.~Lee, S.~Park, and S.~Kim, ``Implementation of trot-to-gallop
  transition and subsequent gallop on the mit cheetah i,'' \emph{The
  International Journal of Robotics Research}, vol.~35, no.~13, pp. 1627--1650,
  2016.

\bibitem{park2017high}
H.-W. Park, P.~M. Wensing, and S.~Kim, ``High-speed bounding with the mit
  cheetah 2: Control design and experiments,'' \emph{The International Journal
  of Robotics Research}, vol.~36, no.~2, pp. 167--192, 2017.

\bibitem{briggs2012tails}
R.~Briggs, J.~Lee, M.~Haberland, and S.~Kim, ``Tails in biomimetic design:
  Analysis, simulation, and experiment,'' in \emph{2012 IEEE/RSJ International
  Conference on Intelligent Robots and Systems}.\hskip 1em plus 0.5em minus
  0.4em\relax IEEE, 2012, pp. 1473--1480.

\bibitem{patel2014rapid}
A.~Patel and M.~Braae, ``Rapid acceleration and braking: Inspirations from the
  cheetah's tail,'' in \emph{2014 IEEE International Conference on Robotics and
  Automation (ICRA)}.\hskip 1em plus 0.5em minus 0.4em\relax IEEE, 2014, pp.
  793--799.

\bibitem{khoramshahi2013piecewise}
M.~Khoramshahi, H.~J. Bidgoly, S.~Shafiee, A.~Asaei, A.~J. Ijspeert, and M.~N.
  Ahmadabadi, ``Piecewise linear spine for speed--energy efficiency trade-off
  in quadruped robots,'' \emph{Robotics and Autonomous Systems}, vol.~61,
  no.~12, pp. 1350--1359, 2013.

\bibitem{eckert2015comparing}
P.~Eckert, A.~Spr{\"o}witz, H.~Witte, and A.~J. Ijspeert, ``Comparing the
  effect of different spine and leg designs for a small bounding quadruped
  robot,'' in \emph{2015 IEEE International Conference on Robotics and
  Automation (ICRA)}.\hskip 1em plus 0.5em minus 0.4em\relax IEEE, 2015, pp.
  3128--3133.

\bibitem{schulman2017proximal}
J.~Schulman, F.~Wolski, P.~Dhariwal, A.~Radford, and O.~Klimov, ``Proximal
  policy optimization algorithms,'' \emph{arXiv preprint arXiv:1707.06347},
  2017.

\bibitem{salimans2017evolution}
T.~Salimans, J.~Ho, X.~Chen, S.~Sidor, and I.~Sutskever, ``Evolution strategies
  as a scalable alternative to reinforcement learning,'' \emph{arXiv preprint
  arXiv:1703.03864}, 2017.

\bibitem{mania2018simple}
H.~Mania, A.~Guy, and B.~Recht, ``Simple random search provides a competitive
  approach to reinforcement learning,'' \emph{arXiv preprint arXiv:1803.07055},
  2018.

\bibitem{47151}
\BIBentryALTinterwordspacing
J.~Tan, T.~Zhang, E.~Coumans, A.~Iscen, Y.~Bai, D.~Hafner, S.~Bohez, and
  V.~Vanhoucke, ``Sim-to-real: Learning agile locomotion for quadruped
  robots,'' in \emph{Robotics: Science and Systems}, 2018. [Online]. Available:
  \url{https://arxiv.org/pdf/1804.10332.pdf}
\BIBentrySTDinterwordspacing

\bibitem{47598}
\BIBentryALTinterwordspacing
A.~Iscen, K.~Caluwaerts, J.~Tan, T.~Zhang, E.~Coumans, V.~Sindhwani, and
  V.~Vanhoucke, ``Policies modulating trajectory generators,'' in \emph{2nd
  Annual Conference on Robot Learning, CoRL 2018}, 2018, pp. 916--926.
  [Online]. Available:
  \url{https://storage.googleapis.com/pub-tools-public-publication-data/pdf/9c90117f324d3b1bc53779713cb1c800841c926a.pdf}
\BIBentrySTDinterwordspacing

\bibitem{hwangbo2019learning}
J.~Hwangbo, J.~Lee, A.~Dosovitskiy, D.~Bellicoso, V.~Tsounis, V.~Koltun, and
  M.~Hutter, ``Learning agile and dynamic motor skills for legged robots,''
  \emph{Science Robotics}, vol.~4, no.~26, p. eaau5872, 2019.

\bibitem{moore2015outrun}
T.~Y. Moore and A.~A. Biewener, ``Outrun or outmaneuver: Predator--prey
  interactions as a model system for integrating biomechanical studies in a
  broader ecological and evolutionary context,'' \emph{Integrative and
  comparative biology}, vol.~55, no.~6, pp. 1188--1197, 2015.

\bibitem{4445757}
L.~{Busoniu}, R.~{Babuska}, and B.~{De Schutter}, ``A comprehensive survey of
  multiagent reinforcement learning,'' \emph{IEEE Transactions on Systems, Man,
  and Cybernetics, Part C (Applications and Reviews)}, vol.~38, no.~2, pp.
  156--172, March 2008.

\bibitem{OpenAI_dota}
OpenAI, ``Openai five,'' \url{https://blog.openai.com/openai-five/}.

\bibitem{alphastarblog}
O.~Vinyals, I.~Babuschkin, J.~Chung, M.~Mathieu, M.~Jaderberg, W.~M. Czarnecki,
  A.~Dudzik, A.~Huang, P.~Georgiev, R.~Powell, T.~Ewalds, D.~Horgan, M.~Kroiss,
  I.~Danihelka, J.~Agapiou, J.~Oh, V.~Dalibard, D.~Choi, L.~Sifre, Y.~Sulsky,
  S.~Vezhnevets, J.~Molloy, T.~Cai, D.~Budden, T.~Paine, C.~Gulcehre, Z.~Wang,
  T.~Pfaff, T.~Pohlen, D.~Yogatama, J.~Cohen, K.~McKinney, O.~Smith, T.~Schaul,
  T.~Lillicrap, C.~Apps, K.~Kavukcuoglu, D.~Hassabis, and D.~Silver,
  ``{AlphaStar: Mastering the Real-Time Strategy Game StarCraft II},''
  \url{https://deepmind.com/blog/alphastar-mastering-real-time-strategy-game-starcraft-ii/},
  2019.

\bibitem{Baker2020Emergent}
\BIBentryALTinterwordspacing
B.~Baker, I.~Kanitscheider, T.~Markov, Y.~Wu, G.~Powell, B.~McGrew, and
  I.~Mordatch, ``Emergent tool use from multi-agent autocurricula,'' in
  \emph{International Conference on Learning Representations}, 2020. [Online].
  Available: \url{https://openreview.net/forum?id=SkxpxJBKwS}
\BIBentrySTDinterwordspacing

\bibitem{44806}
\BIBentryALTinterwordspacing
D.~Silver, A.~Huang, C.~J. Maddison, A.~Guez, L.~Sifre, G.~van~den Driessche,
  J.~Schrittwieser, I.~Antonoglou, V.~Panneershelvam, M.~Lanctot, S.~Dieleman,
  D.~Grewe, J.~Nham, N.~Kalchbrenner, I.~Sutskever, T.~Lillicrap, M.~Leach,
  K.~Kavukcuoglu, T.~Graepel, and D.~Hassabis, ``Mastering the game of go with
  deep neural networks and tree search,'' \emph{Nature}, vol. 529, pp.
  484--503, 2016. [Online]. Available:
  \url{http://www.nature.com/nature/journal/v529/n7587/full/nature16961.html}
\BIBentrySTDinterwordspacing

\bibitem{DBLP:journals/corr/abs-1712-01815}
\BIBentryALTinterwordspacing
D.~Silver, T.~Hubert, J.~Schrittwieser, I.~Antonoglou, M.~Lai, A.~Guez,
  M.~Lanctot, L.~Sifre, D.~Kumaran, T.~Graepel, T.~P. Lillicrap, K.~Simonyan,
  and D.~Hassabis, ``Mastering chess and shogi by self-play with a general
  reinforcement learning algorithm,'' \emph{CoRR}, vol. abs/1712.01815, 2017.
  [Online]. Available: \url{http://arxiv.org/abs/1712.01815}
\BIBentrySTDinterwordspacing

\bibitem{stone1998towards}
P.~Stone and M.~Veloso, ``Towards collaborative and adversarial learning: A
  case study in robotic soccer,'' \emph{International Journal of Human-Computer
  Studies}, vol.~48, no.~1, pp. 83--104, 1998.

\bibitem{45924}
\BIBentryALTinterwordspacing
K.~Bousmalis, N.~Silberman, D.~Dohan, D.~Erhan, and D.~Krishnan, ``Unsupervised
  pixel-level domain adaptation with generative adversarial networks,''
  \emph{CoRR}, vol. abs/1612.05424, 2016. [Online]. Available:
  \url{http://arxiv.org/abs/1612.05424}
\BIBentrySTDinterwordspacing

\bibitem{DBLP:journals/corr/abs-1903-00636}
\BIBentryALTinterwordspacing
J.~Duan, Q.~Wang, L.~Pinto, C.~J. Kuo, and S.~Nikolaidis, ``Robot learning via
  human adversarial games,'' \emph{CoRR}, vol. abs/1903.00636, 2019. [Online].
  Available: \url{http://arxiv.org/abs/1903.00636}
\BIBentrySTDinterwordspacing

\bibitem{pmlr-v70-pinto17a}
\BIBentryALTinterwordspacing
L.~Pinto, J.~Davidson, R.~Sukthankar, and A.~Gupta, ``Robust adversarial
  reinforcement learning,'' in \emph{Proceedings of the 34th International
  Conference on Machine Learning}, ser. Proceedings of Machine Learning
  Research, D.~Precup and Y.~W. Teh, Eds., vol.~70.\hskip 1em plus 0.5em minus
  0.4em\relax International Convention Centre, Sydney, Australia: PMLR, 06--11
  Aug 2017, pp. 2817--2826. [Online]. Available:
  \url{http://proceedings.mlr.press/v70/pinto17a.html}
\BIBentrySTDinterwordspacing

\bibitem{DBLP:journals/corr/LoweWTHAM17}
\BIBentryALTinterwordspacing
R.~Lowe, Y.~Wu, A.~Tamar, J.~Harb, P.~Abbeel, and I.~Mordatch, ``Multi-agent
  actor-critic for mixed cooperative-competitive environments,'' \emph{CoRR},
  vol. abs/1706.02275, 2017. [Online]. Available:
  \url{http://arxiv.org/abs/1706.02275}
\BIBentrySTDinterwordspacing

\bibitem{ackermann2019reducing}
J.~Ackermann, V.~Gabler, T.~Osa, and M.~Sugiyama, ``Reducing overestimation
  bias in multi-agent domains using double centralized critics,'' 2019.

\bibitem{wu2010terrain}
J.-c. Wu and Z.~Popovi{\'c}, ``Terrain-adaptive bipedal locomotion control,''
  \emph{ACM Transactions on Graphics (TOG)}, vol.~29, no.~4, pp. 1--10, 2010.

\bibitem{wang2010optimizing}
J.~M. Wang, D.~J. Fleet, and A.~Hertzmann, ``Optimizing walking controllers for
  uncertain inputs and environments,'' \emph{ACM Transactions on Graphics
  (TOG)}, vol.~29, no.~4, pp. 1--8, 2010.

\bibitem{tan2016simulation}
J.~Tan, Z.~Xie, B.~Boots, and C.~K. Liu, ``Simulation-based design of dynamic
  controllers for humanoid balancing,'' in \emph{2016 IEEE/RSJ International
  Conference on Intelligent Robots and Systems (IROS)}.\hskip 1em plus 0.5em
  minus 0.4em\relax IEEE, 2016, pp. 2729--2736.

\bibitem{Hansen2006}
N.~Hansen, \emph{The CMA Evolution Strategy: A Comparing Review}.\hskip 1em
  plus 0.5em minus 0.4em\relax Berlin, Heidelberg: Springer Berlin Heidelberg,
  2006, pp. 75--102.

\bibitem{DBLP:journals/corr/PengYWYTLW17}
\BIBentryALTinterwordspacing
P.~Peng, Q.~Yuan, Y.~Wen, Y.~Yang, Z.~Tang, H.~Long, and J.~Wang, ``Multiagent
  bidirectionally-coordinated nets for learning to play starcraft combat
  games,'' \emph{CoRR}, vol. abs/1703.10069, 2017. [Online]. Available:
  \url{http://arxiv.org/abs/1703.10069}
\BIBentrySTDinterwordspacing

\bibitem{10.5555/2900929.2901013}
L.~Matignon, L.~Jeanpierre, and A.-I. Mouaddib, ``Coordinated multi-robot
  exploration under communication constraints using decentralized markov
  decision processes,'' in \emph{Proceedings of the Twenty-Sixth AAAI
  Conference on Artificial Intelligence}, ser. AAAI’12.\hskip 1em plus 0.5em
  minus 0.4em\relax AAAI Press, 2012, p. 2017–2023.

\bibitem{DBLP:journals/corr/FoersterAFW16a}
\BIBentryALTinterwordspacing
J.~N. Foerster, Y.~M. Assael, N.~de~Freitas, and S.~Whiteson, ``Learning to
  communicate with deep multi-agent reinforcement learning,'' \emph{CoRR}, vol.
  abs/1605.06676, 2016. [Online]. Available:
  \url{http://arxiv.org/abs/1605.06676}
\BIBentrySTDinterwordspacing

\bibitem{NIPS2016_6398}
\BIBentryALTinterwordspacing
S.~Sukhbaatar, a.~szlam, and R.~Fergus, ``Learning multiagent communication
  with backpropagation,'' in \emph{Advances in Neural Information Processing
  Systems 29}, D.~D. Lee, M.~Sugiyama, U.~V. Luxburg, I.~Guyon, and R.~Garnett,
  Eds.\hskip 1em plus 0.5em minus 0.4em\relax Curran Associates, Inc., 2016,
  pp. 2244--2252. [Online]. Available:
  \url{http://papers.nips.cc/paper/6398-learning-multiagent-communication-with-backpropagation.pdf}
\BIBentrySTDinterwordspacing

\bibitem{DBLP:journals/corr/MordatchA17}
\BIBentryALTinterwordspacing
I.~Mordatch and P.~Abbeel, ``Emergence of grounded compositional language in
  multi-agent populations,'' \emph{CoRR}, vol. abs/1703.04908, 2017. [Online].
  Available: \url{http://arxiv.org/abs/1703.04908}
\BIBentrySTDinterwordspacing

\bibitem{10.5555/1483085}
Y.~Shoham and K.~Leyton-Brown, \emph{Multiagent Systems: Algorithmic,
  Game-Theoretic, and Logical Foundations}.\hskip 1em plus 0.5em minus
  0.4em\relax USA: Cambridge University Press, 2008.

\bibitem{NIPS1999_1786}
\BIBentryALTinterwordspacing
V.~R. Konda and J.~N. Tsitsiklis, ``Actor-critic algorithms,'' in
  \emph{Advances in Neural Information Processing Systems 12}, S.~A. Solla,
  T.~K. Leen, and K.~M\"{u}ller, Eds.\hskip 1em plus 0.5em minus 0.4em\relax
  MIT Press, 2000, pp. 1008--1014. [Online]. Available:
  \url{http://papers.nips.cc/paper/1786-actor-critic-algorithms.pdf}
\BIBentrySTDinterwordspacing

\bibitem{pmlr-v32-silver14}
\BIBentryALTinterwordspacing
D.~Silver, G.~Lever, N.~Heess, T.~Degris, D.~Wierstra, and M.~Riedmiller,
  ``Deterministic policy gradient algorithms,'' in \emph{Proceedings of the
  31st International Conference on Machine Learning}, ser. Proceedings of
  Machine Learning Research, E.~P. Xing and T.~Jebara, Eds., vol.~32,
  no.~1.\hskip 1em plus 0.5em minus 0.4em\relax Bejing, China: PMLR, 22--24 Jun
  2014, pp. 387--395. [Online]. Available:
  \url{http://proceedings.mlr.press/v32/silver14.html}
\BIBentrySTDinterwordspacing

\bibitem{DBLP:journals/corr/abs-1802-09477}
\BIBentryALTinterwordspacing
S.~Fujimoto, H.~van Hoof, and D.~Meger, ``Addressing function approximation
  error in actor-critic methods,'' \emph{CoRR}, vol. abs/1802.09477, 2018.
  [Online]. Available: \url{http://arxiv.org/abs/1802.09477}
\BIBentrySTDinterwordspacing

\bibitem{DBLP:journals/corr/FoersterFANW17}
\BIBentryALTinterwordspacing
J.~N. Foerster, G.~Farquhar, T.~Afouras, N.~Nardelli, and S.~Whiteson,
  ``Counterfactual multi-agent policy gradients,'' \emph{CoRR}, vol.
  abs/1705.08926, 2017. [Online]. Available:
  \url{http://arxiv.org/abs/1705.08926}
\BIBentrySTDinterwordspacing

\bibitem{DBLP:journals/corr/abs-1710-03748}
\BIBentryALTinterwordspacing
T.~Bansal, J.~Pachocki, S.~Sidor, I.~Sutskever, and I.~Mordatch, ``Emergent
  complexity via multi-agent competition,'' \emph{CoRR}, vol. abs/1710.03748,
  2017. [Online]. Available: \url{http://arxiv.org/abs/1710.03748}
\BIBentrySTDinterwordspacing

\bibitem{Chen:2016:LML:3086758}
Z.~Chen and B.~Liu, \emph{Lifelong Machine Learning}.\hskip 1em plus 0.5em
  minus 0.4em\relax Morgan \& Claypool Publishers, 2016.

\bibitem{yu2018learning}
W.~Yu, G.~Turk, and C.~K. Liu, ``Learning symmetric and low-energy
  locomotion,'' \emph{ACM Transactions on Graphics (TOG)}, vol.~37, no.~4, pp.
  1--12, 2018.

\bibitem{laikago-intro}
E.~Guizzo, ``This robotics startup wants to be the boston dynamics of china,''
  \url{https://spectrum.ieee.org/automaton/robotics/robotics-hardware/this-robotics-startup-wants-to-be-the-boston-dynamics-of-china},
  2017.

\bibitem{coumans2020}
E.~Coumans and Y.~Bai, ``Pybullet, a python module for physics simulation for
  games, robotics and machine learning,'' \url{http://pybullet.org},
  2016--2020.

\bibitem{hansen2019pycma}
\BIBentryALTinterwordspacing
N.~Hansen, Y.~Akimoto, and P.~Baudis, ``{CMA-ES/pycma} on {G}ithub,'' Zenodo,
  DOI:10.5281/zenodo.2559634, Feb. 2019. [Online]. Available:
  \url{https://doi.org/10.5281/zenodo.2559634}
\BIBentrySTDinterwordspacing

\bibitem{wilson2013locomotion}
A.~M. Wilson, J.~Lowe, K.~Roskilly, P.~E. Hudson, K.~Golabek, and J.~McNutt,
  ``Locomotion dynamics of hunting in wild cheetahs,'' \emph{Nature}, vol. 498,
  no. 7453, p. 185, 2013.

\bibitem{RoboImitationPeng20}
X.~B. Peng, E.~Coumans, T.~Zhang, T.-W. Lee, J.~Tan, and S.~Levine, ``Learning
  agile robotic locomotion skills by imitating animals,'' 2020.

\end{thebibliography}

\end{document}